\pdfoutput=1

\documentclass[11pt]{article}

\usepackage[final]{acl}

\usepackage{times}
\usepackage{latexsym}

\usepackage[T1]{fontenc}

\usepackage[utf8]{inputenc}

\usepackage{microtype}

\usepackage{inconsolata}

\usepackage{graphicx}

\title{Improving Hate Speech Classification with Cross-Taxonomy \\Dataset Integration}

\author{Jan Fillies\textsuperscript{1,2} \and \textbf{Adrian Paschke}\textsuperscript{1,2,3} \\
         $^{1}$Institut für Angewandte Informatik, Leipzig, Germany \\
         $^{2}$Freie Universität Berlin, Berlin, Germany \\
         $^{3}$Fraunhofer-Institut für Offene Kommunikationssysteme, Berlin, Germany \\
         \texttt{jan.fillies@fu-berlin.de}}

\begin{document}
\maketitle
\begin{abstract}
Algorithmic hate speech detection faces significant challenges due to the diverse definitions and datasets used in research and practice. Social media platforms, legal frameworks, and institutions each apply distinct yet overlapping definitions, complicating classification efforts. This study addresses these challenges by demonstrating that existing datasets and taxonomies can be integrated into a unified model, enhancing prediction performance and reducing reliance on multiple specialized classifiers. The work introduces a universal taxonomy and a hate speech classifier capable of detecting a wide range of definitions within a single framework. Our approach is validated by combining two widely used but differently annotated datasets, showing improved classification performance on an independent test set. This work highlights the potential of dataset and taxonomy integration in advancing hate speech detection, increasing efficiency, and ensuring broader applicability across contexts.
\end{abstract}

\section{Introduction}
\label{Introduction}
Research has shown a direct link between the rise of online hate speech and offline events \cite{Lupu2023}, highlighting the growing impact of digital platforms on real-world occurrences. As of April 2023, there are an estimated 4.8 billion global social media users, making up about 59.9\% of the world’s population \cite{Kemp_2023}. This massive reach underscores the scale of the problem, with Facebook alone removing 38.3 million instances of hate speech in the first three quarters of 2023 \cite{statis_2023}. These numbers emphasize both the urgency and magnitude of the issue, making it a top priority for the research community. The challenge lies in balancing the preservation of free speech with the need to protect individuals from harm. While algorithms play a key role in addressing this issue, they are just one part of a broader, multi-faceted approach. In this context, this research aims to develop efficient and effective algorithmic solutions for hate speech detection.

One main challenge in the field is that the understanding of hate speech varies and is influenced by factors such as topic \citep{Wiegand2019}, author \citep{nejadgholi-kiritchenko-2020-cross}, and time \citep{justen2022}, among others. Even within the legal context, it is a complex process deciding whether a statement should be classified as hateful or not. In response, research, private, and public entities have developed their own definitions and community standards, legal frameworks, or annotation guidelines \citep{macavaney2019hate}. \par
Especially in the research field, the available datasets heavily depend on the annotation procedure and the definitions of hate speech provided to the annotators \cite{vidgen2020directions}. This dependence and wide variety of definitions  makes it challenging to compare \cite{Fortuna2018} or merge datasets annotated within different annotation schemas. While the field of available annotated hate speech corpora is limited to begin with, this additional limitation of incompatibility further complicates efforts to provide general and reliable hate speech detection. \par
This research addresses this gap by providing a machine learning structure that combines existing definitions and datasets. It identifies mismatches in definitions, faults during the annotation combining process, and missing labels in datasets. The study demonstrates the feasibility of merging annotation schemas and datasets to detect a wider variety of hate speech definitions using just one trained classifier. It establishes that a single general taxonomy can be created and employed for multi-label federated training of a classifier, thereby improving prediction quality.\par
The approach is evaluated using two standard research datasets and their respective definitions. The outcome involves the creation of a comprehensive hate speech taxonomy and the training of a general hate speech classifier. \par
The scripts used for preprocessing, dataset construction, training, and evaluation are available as part of the paper.\footnote{https://github.com/fillies/HateSpeechCrossTaxonomy\\DatasetIntegration} This offers a deeper insight and facilitates the reproducibility of our work. Please note that the used datasets have to be obtained from the cited sources. 


\section{Related Work}
\label{relatedwork}
\textbf{Datasets} -   The field of hate speech datasets is rapidly growing. Established datasets include \citep{Hosseinmardi2015, Gibert2018, ElSherief2018}, while newer, smaller datasets \cite{fillies2023hateful,fillies2025novel} continue to emerge. A comprehensive overview is provided by \citet{vidgen2020directions}. Analysis of these datasets highlights diverse annotation schemes \citep{Chung2019}, from binary labels to multi-class hierarchies \citep{Ranasinghe2020}. Universal annotation frameworks are also recognized \citep{Bartalesi2006}. However, no single benchmark dataset or universally accepted definition of hate speech exists \citep{macavaney2019hate}. The wide range of definitions has been extensively studied by \citet{stephan2020}.


\textbf{Algorithmic Detection} - For detecting hate speech, toxic speech, abusive language, and related areas, the predominant algorithmic approach has utilized supervised transformer-based architectures \cite{mozafari2020bert, poletto2021resources, plaza2023respectful}. Fine-tuning transformer models, particularly BERT \cite{devlin2019bert}, has demonstrated significant performance enhancements compared to other methods \cite{liu2019nuli, kirk2022handling, fillies2023multilingual}. Recently, the focus has shifted towards using pre-trained large language models combined with prompting techniques for hate speech detection \cite{kim2023conprompt, plaza2023respectful, fillies2024simple}.

\textbf{Taxonomy and Ontology Matching} - Several researchers have aimed to create general hate speech ontologies \citep{stranisci-etal-2022-o, sharma-etal-2018-degree} and taxonomies \citep{Salminen2018, zufall-etal-2022-legal, lewandowska2023llod}. \citet{Salminen2018} integrated their taxonomy into a transformer-based hate speech detection model, partially building on existing taxonomies and combining them to annotate a new dataset. The practice of merging ontologies is well established \citep{Shvaiko2013}. However, no research has yet combined hate speech taxonomies to make existing datasets suitable for iterative federated learning.\par
\textbf{Federated and Continuous Learning} - Federated learning for hate speech detection is crucial as it mitigates privacy concerns related to data sharing. A key development is \citet{zampieri2024federated}, which introduces a binary hate speech classifier using a decentralized architecture, demonstrating superior performance across datasets while preserving privacy. Another significant study, \citet{gala2023federated}, explores multi-class federated learning on a static dataset with uniform annotations, disregarding annotation mismatches and emphasizing distributed training benefits. In continuous learning, \citet{omrani2023towards} propose a novel framework for detecting problematic content by integrating various datasets and treating each label as an independent classification task.\par

This research directly builds upon the work of \citet{zampieri2024federated}, \citet{gala2023federated}, and \citet{omrani2023towards}. It extends the findings of \citet{zampieri2024federated} and \citet{gala2023federated} by demonstrating that federated training for hate speech detection is feasible not only for binary classification but also for multi-label hate speech datasets with varying definitions of hate speech. In relation to \citet{omrani2023towards}, it advances the research by integrating labels into a unified taxonomy with hierarchical aspects, introducing a deeper semantic relationship model, and showing that this model can be continuously adapted. \par

\section{Methodology}
\label{methodology}
\begin{figure}[ht]
  \centering
  \includegraphics[width=\linewidth]{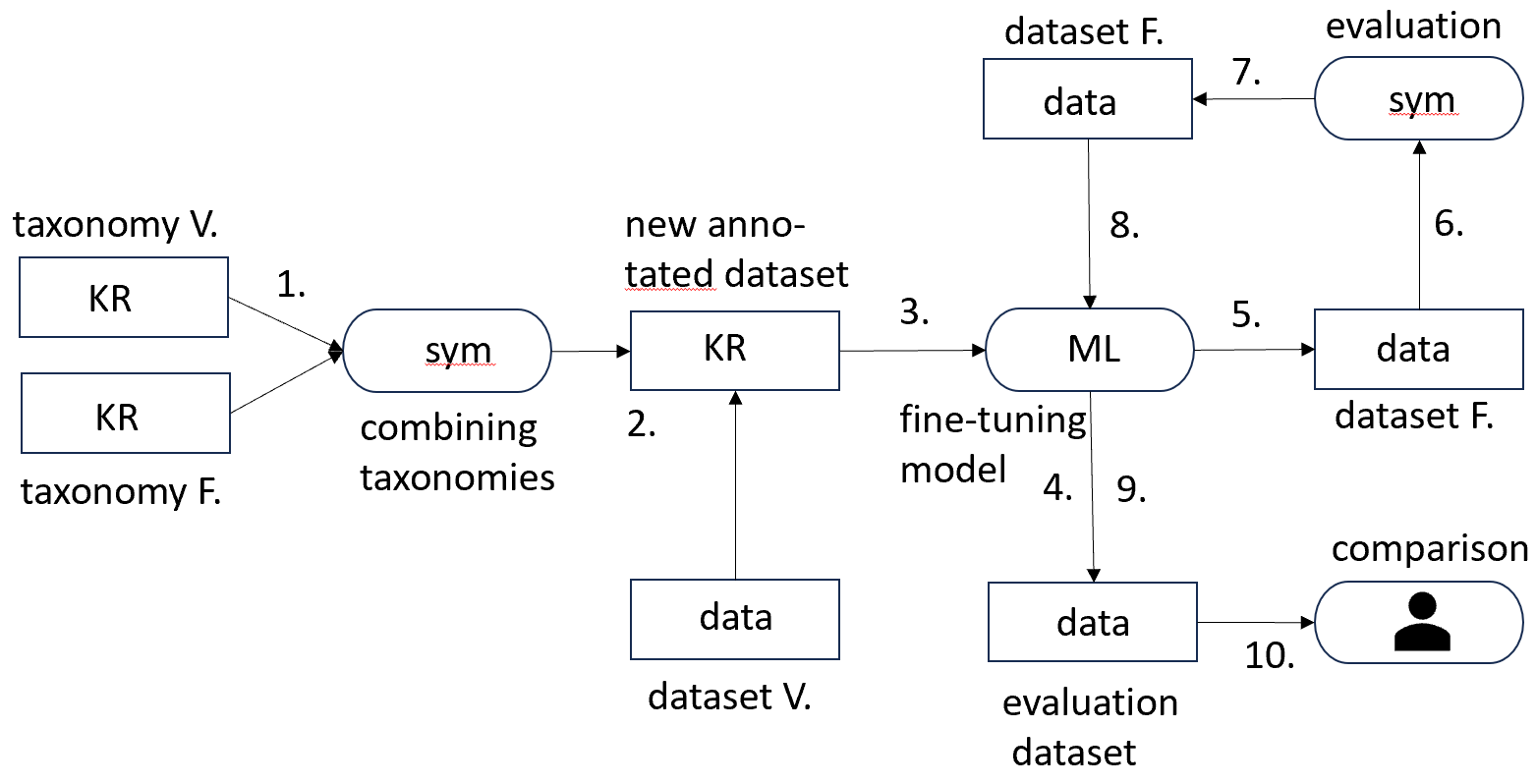}
  \caption{Boxology-Model of the Process. F. = Fanton et al. 2021,  V. = Vidgen et al. 2021, sym = Symbolic Processing KR. = Knowledge Representation, ML = Machine Learning }
  \label{Boxology}
\end{figure}

The research is divided into three main parts. First, a general hate speech taxonomy is created. Second, this taxonomy is used to fine-tune a pre-trained multi-label hate speech detection model multiple times on two different datasets (see Sections \ref{dataset} and \ref{experimentalclassifier}). Lastly, continuous evaluation is conducted after each training cycle. Each step is detailed in this section (see Figure \ref{Boxology}), with selected datasets, taxonomies, and the  models serving as examples to demonstrate the approach's functionality.

\begin{enumerate}
    \item In the first step, the taxonomies are combined into one general taxonomy. Here, the general taxonomy should include all the classes proposed by the underlying concepts. A class hierarchy is introduced to represent and adjust to different levels of abstraction (see section \ref{generaltax}). In this step, classes that cannot be merged are identified and removed. A word-level matching of annotations between the original and the new general taxonomy is introduced. The class hierarchy of the general taxonomy is represented through a one-hot encoded vector; when a subclass is flagged as identified, the parent classes must be present too.
    \item In the following step, one dataset is selected to have its annotations transferred into the new annotation format based on the general taxonomy. Here, it is expected that certain flags within the annotations are missing or, more precisely, incorrectly annotated.
    \item Based on this newly annotated dataset, a multi-label classifier is trained (see section \ref{experimentalclassifier}).
    \item To validate the performance of the trained model and provide insight into the generalizability of the model, an  external binary hate speech dataset is provided as an evaluation dataset, and the performance is measured (see section \ref{results}).
    \item The trained classifier is now used to predict all known labels of the second dataset.
    \item The True Positive, False Negative, False Positive, and True Negative distributions of the predictions generate insights into three main aspects regarding the annotations. Firstly, it can be observed where the definitions of concepts are not aligned. Secondly, it can be determined if the general taxonomy made a mistake in its hierarchical structure. Lastly, it can be identified which flags are not represented in the old annotation of the new dataset (see section \ref{disscussionofresults}).
    \item After evaluation, the prediction scores and the human annotations of the second dataset can be combined. In the parts where the human annotation identified a hateful instance, they overwrite the given predictions. Classes that had to be excluded due to definition mismatches can be annotated, but only with the predictions of the network. The predicted values are normalized to [0,1], while the human annotations remain binary.
    \item Based on this mix of predicted and human-based annotations, the original network is fine-tuned again on the new dataset (see section \ref{results}). Extra measures to prevent overfitting can be implemented.
    \item The dataset is evaluated again using the same binary hate/no-hate external dataset (see section \ref{results}).
    \item Lastly, the two measurements of prediction quality on the external dataset are compared to validate the performance and provide insight into generalizability (see section \ref{disscussionofresults}).
\end{enumerate}

\section{Datasets}
\label{dataset}
Two primary datasets with different annotations were selected for this research, along with two additional datasets: one for evaluation and one for balancing the two main datasets during training with non-hateful statements.

The first main dataset, provided by \citet{vidgen-etal-2021-learning}, is a large, dynamically generated collection of 41,255 entries created over four rounds, with 54\% of the entries being hateful. The dataset includes 11 English-language training datasets for hate and toxicity from hatespeechdata.com. Its hierarchical taxonomy, based on \citet{Nickerson2013}, classifies entries into hate and no-hate categories. The hate entries are further divided into five types (Derogation, Animosity, Threatening Language, Support for Hateful Entities, Dehumanization). Additionally, 29 identities as hate targets are annotated. The annotations were performed by 20 trained annotators.


The second main dataset compiled by \citet{fanton2021} is also a dynamically generated human-in-the-loop dataset, containing 5,000 hateful statements. Created over two cycles with human input in between, the initial dataset included 880 statements and was developed in collaboration with 20 experts from various NGOs. The annotations featured 10 labels (“DISABLED,” “JEWS,” “OVERWEIGHT,” “LGBT+,” “MUSLIM,” “WOMEN,” “PEOPLE OF COLOR,” “ROMANI,” “MIGRANTS,” “OTHER”). Three trained students were involved in the annotation process.

The dataset from \citet{fillies2023hateful} was selected for non-hateful statements, as only the hateful entries were selected from the two main datasets, and training a classifier solely on those would likely result in overfitting. This dataset, in English, includes annotated Discord messages collected between March 2021 and June 2022, comprising 88,395 chat messages. Around 6.42\% of the messages were classified as hate speech.

The final support dataset, from \citet{ljube2021}, was chosen for validation and independent evaluation of the classifier's performance. It consists of YouTube comments collected between January and May 2020, with approximately 50\% hate and 50\% non-hateful examples.


\section{General Taxonomy}
\label{generaltax}

\begin{figure}[ht]
  \centering
  \includegraphics[width=3in]{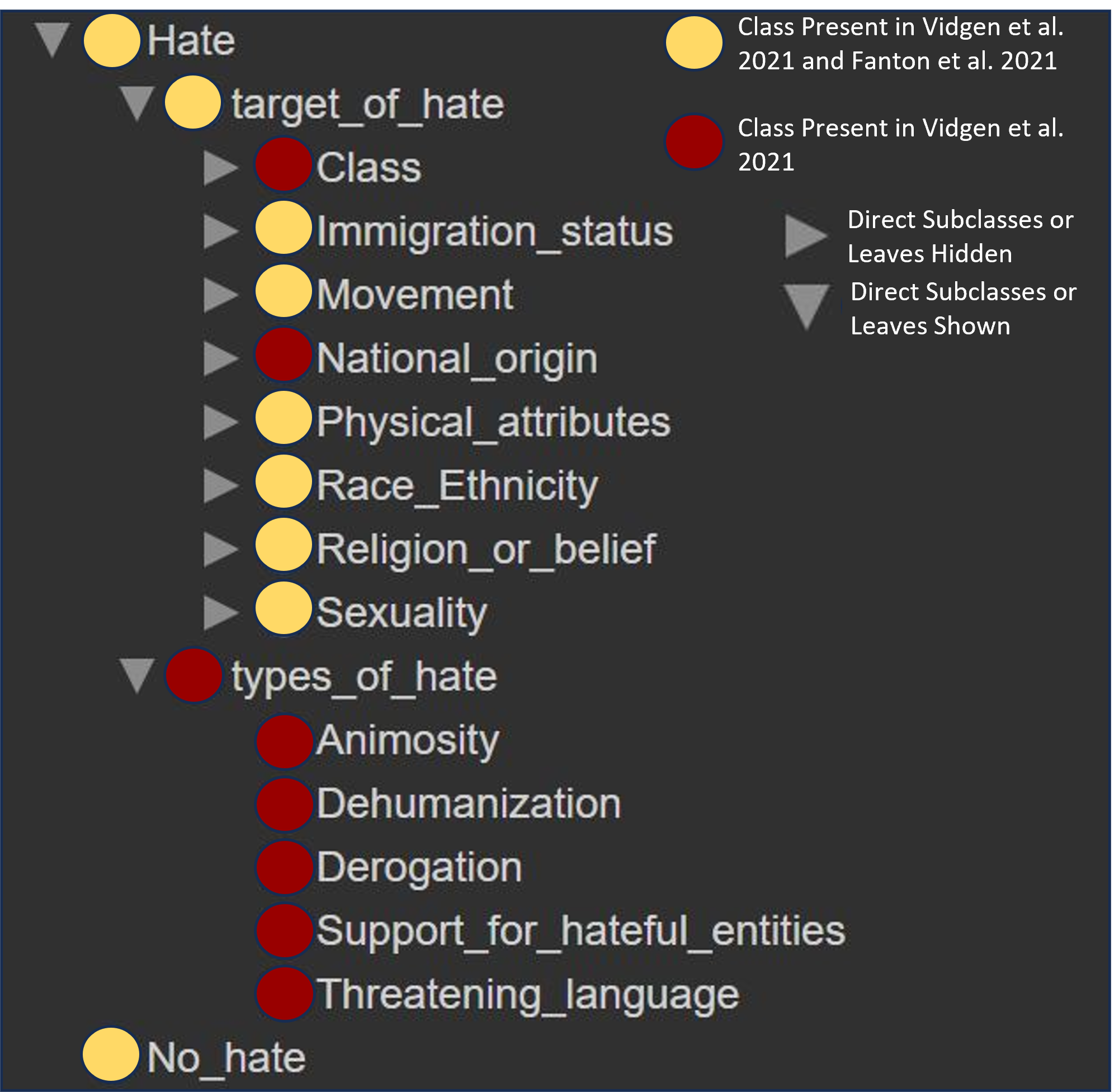}
  \caption{Overview General Taxonomy Level 1 - 3}
  \label{generlatax_extract}
\end{figure}

This research explores merging multiple taxonomies into a central one to enable a single classifier to predict diverse definitions using differently annotated datasets. As a demonstration, two existing taxonomies were combined. The taxonomy was developed by a two-person team and is shown in Appendix \ref{Generaltaxonomy}, with the first three levels in Figure \ref{generlatax_extract}. Shared classes and leaves (labels not further broken down) are highlighted in yellow, while those unique to \citet{vidgen-etal-2021-learning} are in red. Both taxonomies contributed different, identical, or new subclasses and leaves. The final taxonomy has five layers.

The taxonomy from \citet{vidgen-etal-2021-learning} formed the basis for the merge due to its thoroughness. It initially distinguishes between hate and non-hate statements.

Hate types from \citet{vidgen-etal-2021-learning} were grouped under the label "types\_of\_hate," which was absent in \citet{fanton2021}. Adjustments were made for hate targets, with seven out of 11 classes from \citet{fanton2021} fitting directly into the new taxonomy. The remaining classes, like “Gender,” “Intersectional,” and “Disability,” required modifications.

Due to \citet{fanton2021} introducing the labels “Disabled” and “Overweight”, a class regarding physical attributes was introduced, also containing the label “Gender,” which then includes the class “Gender Minorities,” unlike \citet{vidgen-etal-2021-learning} where it is independent. The last label from \citet{vidgen-etal-2021-learning}, “Intersectional,” was not included explicitly, as it is contained in the multi-label encodings (e.g., black women) that are represented in the taxonomy.

The classes (“Jews”, “Muslim”, “Women”, “Romani”, “Migrants”) from \citet{fanton2021} were already covered in the taxonomy. The label “People of color” from \citet{fanton2021} was initially introduced as an independent label under the class “Physical\_attributes/skin\_color/” next to the labels “Black” and “White.” However, the evaluation of the trained network's performance clearly showed this as a mistake, making it necessary to make “Black” a subclass of “People of Color.”

The main challenge was the label “LGBT+” by \citet{fanton2021} due to its covering of multiple aspects. It is first a political and social movement, standing for “lesbian, gay, bisexual, transgender, plus other sexual and gender identities,” making it difficult to locate in the existing classes of gender and sexual orientation. The decision was made to include it in the taxonomy as a movement.

It is noteworthy that in the actual dataset annotations by \citet{vidgen-etal-2021-learning}, labels appeared that were not represented in the provided taxonomy, such as “old.people,” “russian,” “lgbtq,” “eastern.europe,” and “non.white.” These labels were included in the new general taxonomy with their own classes. However, the label “other” from \citet{fanton2021} had to be disregarded.
The final taxonomy consists of 23 classes and 43 leaves, merging labels from both taxonomies directly or through abstraction.



\section{Experimental Classifier}
\label{experimentalclassifier}
This section describes the creation of an experimental classifier. The classifier proves the validity of the concept as a proof-of-work. As detailed in the methodology section (\ref{methodology}), the labels in the existing datasets from \citet{vidgen-etal-2021-learning} can be matched to the labels of the new taxonomy, creating a new annotation schema for the dataset. The annotated dataset is then used to fine-tune a pretrained language model to be a multi-label hate speech classifier. After this initial training, the classifier is used to reannotate the second dataset from \citet{fanton2021}, introducing the new annotation schema and providing insights into the created taxonomy, missing labels, and different underlying definitions of hate contained in the two datasets.

The predicted annotations can then be merged with the existing human annotations and used to fine-tune the network again. If the approach holds merit, the minimum requirement is that the hate speech prediction quality of the network increases on an independent test set after the training cycles. This section describes the steps of this process.

\subsection{Encoding}
\label{encoding}
The goal is to map the taxonomy into a network-readable format while preserving class structure information and enabling the annotation of multiple definitions within a unified schema. The proposed encoding uses a sparse binary vector, where each position corresponds to a class or leaf in the taxonomy. This allows the network to learn parent-child relationships while capturing varying degrees of hate within a single framework.

For example, in the schema “Target\_of\_hate / Physical\_attributes / Skin\_color / People\_of\_color / Black,” a statement expressing hate toward Black people would be encoded as [1,1,1,1,1], while hate toward people of color would be [1,1,1,1,0]. This approach enables the network to recognize hierarchical relationships and adapt to different depths of hate speech definitions.


\subsection{Evaluation Metrics}
\label{evaluationmetrics}
Two evaluation metrics were used: accuracy and F-1 scores. For a deeper understanding of the results, the distributions of predictions in regard to the human-annotated labels were evaluated in the four groups: True Positive (TP), True Negative (TN), False Positive (FP), and False Negative (FN).

Accuracy is defined as the ratio of correct predictions to the number of total predictions. The F1-Score metric is beneficial in situations where datasets have imbalanced class distributions \citep{tsourakis2022machine}, fitting the problem at hand. For the F-1 Score, a threshold of 0.5 was chosen. 

\subsection{Algorithm}
\label{algorithm}
\normalsize 
As a base, the state-of-the-art model RoBERTa was chosen, first introduced by \citet{liu2019roberta}. It is a fine-tuned, improved version of the BERT model pretrained and introduced by \citet{devlin2019bert}. RoBERTa uses the same architecture as BERT but applies a different tokenizer and pretraining scheme. The research used the pretrained multi-label RoBERTa model for multi-label sequence classification provided through the platform HuggingFace.\footnote{https://huggingface.co/docs/transformers/model\_doc /roberta\#transformers.
RobertaForSequenceClassification} In combination with the fitting tokenizer from “twitter-roberta-base-emotion”.\footnote{https://huggingface.co/cardiffnlp}. It is meant to be an example implementation to show merit.

\subsection{Technical Setup}
\label{techncilsetup}
For training, Google Colaboratory (Colab) was used, providing a browser-based environment for writing and executing Python code in Jupyter notebooks. As noted by \citet{Kimm2021Performance}, Colab offers access to TPUs and GPUs without requiring additional configuration. For all training sessions, a cluster with Nvidia V100 GPUs, 12.7 GB System-RAM, 16 GB GPU-RAM, and 72.8 GB Storage was utilized. The first training cycle took 45 minutes, while the second cycle took 5 minutes. For both cycles, a fixed seed was used, with the evaluation step size set to 500, train and evaluation batch sizes set to 6, and the number of training epochs set to 4. Other hyperparameters followed the default recommendations from RoBERTa. To prevent overfitting during the second cycle, the dropout ratio for attention probabilities and the dropout probability for fully connected layers in embeddings, encoder, and pooler were both set to 0.5.

\subsection{Data preparation}
\label{datapreperation}
As detailed in Section \ref{encoding}, both datasets were encoded using sparse one-hot encoding based on the taxonomy. They were cleaned of duplicates, missing data, and unusable annotations. Given the nature of BERT models, no additional text preprocessing was performed to preserve information. Since both datasets lacked non-hateful language, 30\% non-hateful statements from \citet{fillies2023hateful} were randomly added. Considering that only ~6\% of the 88,000 messages in \citet{fillies2023hateful} contain hate, the risk of including complex cases like counter-hate speech was minimal. These non-hateful examples were also one-hot encoded. A 10\% holdout set was reserved for evaluation, and both datasets were randomized.

After cleaning and adding 30\% non-hate speech statements, the dataset from \citet{vidgen-etal-2021-learning} contained 18,380 instances, while the dataset based on \citet{fanton2021} had 4,767 instances.



The annotation of the \citet{fanton2021} dataset combined human annotations from \citet{fanton2021} with predictions from the first training cycle. When the network failed to predict a label but an annotator identified it, the human annotation took precedence. This approach is justified, as human annotations rely on inter-annotator agreement, reducing the likelihood of false positives, since multiple annotators would need to select the same incorrect label. When no human labels were available or the human annotation didn't match the network's prediction, the network's predictions were used. This was necessary because certain labels were not annotated in the second dataset, and false negatives by annotators were more likely, given that inter-annotator agreement was reduced to binary decisions. For example, the network might predict a low likelihood of racism in a statement (e.g., a score of 0.2 on a scale from -1 to 1). However, human annotation, based on a binary majority agreement among three annotators (two say no racism detected, but one identifies racism), could be flawed. In such cases, the network's prediction is considered a more accurate reflection of reality than the potentially flawed binary annotation.



\begin{table}[ht]
\caption{All Training and Evaluation Test Set Results }
\label{tab:all}
\centering

\begin{tabular}{llrr} 
\hline
        \textbf{Cycle} & \textbf{Dataset}& \textbf{F1-Score}&\textbf{Accuracy}  \\
\hline
        Cycle-1	&Vidgen&	0.89	&0.46\\
        Cycle-1	&ETS&0.73&	-\\
        Cycle-1-A&	Vidgen&	0.89&	0.55\\
        Cycle-1-A&	ETS	&0.73	&-\\
        Cycle-2	&Fanton&	0.91&	0.74\\
        Cycle-2&	ETS	&0.84	&-\\
\hline
\end{tabular}
\end{table}

\begin{table}
\caption{Display of selected classes from the class wise prediction's evaluation of RoBERTa-Cycle-1 on the dataset by \protect\citep{fanton2021}}
\label{tab:selecetdclasses_cycle1}
    \begin{tabular}{lrr}
        \hline
        \textbf{Class/Leaves} &\textbf{F1-Score}& \textbf{Instances}\\
            \hline
            Hate&   	1.00      &	3539\\
            Target\_of\_hate          &	0.99     & 	3539\\
            Movement    &   	0.00   &   	465\\
            LGBTQ+     & 	0.00      &	465\\
            Physical\_attri     &   	0.90      &	1036\\
            Skin\_color    &  	0.93     &  	301\\
            Black     &     	0.00   &   	0\\
            Non\_white       & 	0.03    &   	301\\
            Religion/belief  &   	0.99  &    	1401\\
            Jews       	&0.99 &      	418\\
            Muslims   &   	0.98  &     	983\\
            Sexuality  &    	0.00      &	0\\
            Bisexual &    	0.00    &  	0\\
            Gay     &   	0.00  &    	0\\
            Types\_of\_hate      &    	0.00     & 	0\\
            Weighted avg       	     &	0.89   &  	15017\\
        \hline
    \end{tabular}
\end{table}

\begin{table}[ht]
\caption{Display of selected classes where the class "LGBTQ+" gets miss labeled to. Using the RoBERTa- Cycle-1 model on the dataset by \protect\citet{fanton2021}}
\label{tab:percentagemisslabeled}
\centering
\begin{tabular}{lr} 
\hline
\textbf{Class} & \textbf{Percentage}   \\
\hline
        Physical\_attri.&	0.308\\
        Gender	&0.295\\
        Gender\_min.	&0.189\\
        Trans&	0.166\\
        Women&	0.037\\
        Sexuality	&0.850\\
        Gay&	0.819\\
\hline
\end{tabular}
\end{table}

\begin{table}[ht]
\caption{Display of selected classes where "Non\_white" is mislabeled, using the RoBERTa-Cycle-1 model on the dataset by \citep{fanton2021}.}
\label{tab:percentagemisslabelednonwhite}
\centering
\begin{tabular}{lr} 
\hline
\textbf{Class} & \textbf{Percentage}   \\
\hline
        Black	&0.882 \\
        Race\_Ethnicity&	0.078 \\
\hline
\end{tabular}
\end{table}

\subsection{Results}
\label{results}
The prediction results from the three fine-tuning experiments and their evaluation on the independent evaluation test set (ETS) are shown in Table \ref{tab:all}. The details of these results are discussed individually in section \ref{disscussionofresults}.

\subsubsection{RoBERTa-Cycle-1}
In the first stage, the classifier (RoBERTa-Cycle-1) was trained on the dataset from \citet{vidgen-etal-2021-learning} and evaluated on the evaluation dataset from \citet{ljube2021}.

This training and evaluation were followed by an analysis of the classifier’s predictions at the class level for the dataset from \citet{fanton2021} (see Table \ref{tab:selecetdclasses_cycle1}). For each class, results were assessed, and performance drops, such as in the cases of ‘Non\_white’ and ‘LGBTQ+’, were identified. Incorrectly associated labels were pinpointed (see Table \ref{tab:percentagemisslabeled} and \ref{tab:percentagemisslabelednonwhite}). For instance, many statements labeled ‘LGBTQ+’ were misclassified under the "Sexuality" label. Table \ref{tab:percentagemisslabeled} shows the percentages of other classes predicted for the "LGBTQ+" label, while Table \ref{tab:percentagemisslabelednonwhite} shows the misclassification for "Non\_white". The percentages do not add up to 1, as this is a multi-label prediction with binary annotations. 

These misclassifications highlight the need for adjustments in the taxonomy, as "LGBTQ+" and "Non\_white" are not correctly represented. This led to the need to relabel and retrain the model, resulting in RoBERTa-Cycle-1-A.

\subsubsection{RoBERTa-Cycle-1-A}
In the following, the model RoBERTa-Cycle-1-A and its performance on the Evaluation Test were established, see Table \ref{tab:all}. It can be observed that the F-1 score remains stable while the accuracy increases significantly after adjusting the taxonomy. All prediction results for all classes of the datasets can be found on GitHub\footnote{https://github.com/fillies/HateSpeechCrossTaxonomy\\DatasetIntegration}. 
Table \ref{tab:selecetdclasses_cycle1A} displays a selection of classes important for evaluating the adjustment of the taxonomy in the previous step. 

After the training of RoBERTa-Cycle-1-A, the same in-depth evaluation of the classifier’s predictions on a class level for the dataset from \citet{fanton2021} was performed, see GitHub\footnote{https://github.com/fillies/HateSpeechCrossTaxonomy\\DatasetIntegration}.
This time, no outlier class, in terms of prediction performance, was identified, indicating that there is no further need for adjustment.

\begin{table}[ht]
\caption{Display of selected classes from the class wise predictions evaluation of RoBERTa- Cycle-1-A on the dataset by \protect\citep{fanton2021}}
\label{tab:selecetdclasses_cycle1A}
\centering
\begin{tabular}{lrr} 
\hline
\textbf{Class/Leaves} &\textbf{F1-Score}& \textbf{Instances} \\
\hline
        Hate	  &    	1.00   &   	3539\\
        Target\_of\_hate   &   	0.99     & 	3539\\
        Skin\_color  & 	0.94     &  	301\\
        Non\_white&0.94  &	301\\
        Black      &  	0.00    &  	0\\
        Weighted avg       	   &	0.91   &  	-\\
\hline
\end{tabular}
\end{table}

\subsubsection{RoBERTa-Cycle-2}
Based on RoBERTa-Cycle-1-A and the merged machine and human annotations of the \citet{fanton2021} dataset, the model RoBERTa-Cycle-2 was trained and evaluated on the Evaluation Test Set, see Table \ref{tab:all}. A relevant increase in F1-Score (from 0.73 to 0.84) on the ETS can be observed, accompanied by a general increase in prediction quality on the new dataset (to a new F1-Score of 0.91 and an accuracy of 0.74).

Different from RoBERTa-Cycle-1 and similar to RoBERTa-Cycle-1-A, the evaluation of each annotated class and its prediction performance, see Table \ref{tab:selecetdclasses_cycle2}, did not produce noteworthy outliers in regard to underperformance. Therefore, no further adjustment of the taxonomy is necessary. All prediction results for all classes across all datasets can be found on GitHub\footnote{https://github.com/fillies/HateSpeechCrossTaxonomy\\DatasetIntegration}.

\begin{table}[ht]
\caption{Display of selected classes from the class wise predictions evaluation of RoBERTa-Cycle-2 on the dataset from \protect\citet{vidgen-etal-2021-learning}}
    \label{tab:selecetdclasses_cycle2}
\centering

\begin{tabular}{lrr}
\hline
\textbf{Class/Leaves} &\textbf{F1-Score}& \textbf{Instances} \\
\hline
        Hate&	 	1.00    &  	14900\\
        Target\_of\_hate       & 	1.00    &  	14780\\
        Movement &  	0.00    &  	0\\
        LGBTQ+    & 	0.00     & 	0\\
        Physical\_attributes     &   	0.93      &	7541\\
        Skin\_color  &  	0.88     &  	2918\\
        Black      & 	0.86    &  	2553\\
        Non\_white   &    	0.89  &     	2918\\
        Religion\_or\_belief    & 	0.86     & 	2529\\
        Jews    &   	0.87     & 	1293\\
        Muslims     &  	0.84    &   	1267\\
        Sexuality      & 	0.89 &	1552\\
        Bisexual      &	0.00     & 	110\\
        Gay     &	0.87   &   	1487\\
        Types\_of\_hate        & 	1.00     & 	14900\\
        Weighted avg      &	0.82    & 	-\\
\hline
\end{tabular}
\end{table}

\subsection{Discussion of Results}
\label{disscussionofresults}
\subsubsection{RoBERTa-Cycle-1}
After the first training cycle on the dataset from \citet{vidgen-etal-2021-learning}, the results in table \ref{tab:all}, particularly the F1-Score, show strong performance for the RoBERTa-Cycle-1 classifier. The notable difference between F1-Score and Accuracy highlights the class imbalance, which corresponds with the sparse input vectors and unbalanced class distributions in the dataset. The F1-Score of 0.73 on the Evaluation Test Set further confirms that the classifier successfully learned and generalized the key aspects of hate speech.


The predictions from RoBERTa-Cycle-1 on the \citet{fanton2021} dataset (see Table \ref{tab:selecetdclasses_cycle1}) show that the model excels at identifying higher levels of abstraction, especially in binary hate speech classification, but struggles with more specific categories. Three issues are observed. First, annotations, such as "types\_of\_hate," are missing from the \citet{fanton2021} annotations. 

Second, while the network performs well in predicting the "skin\_color" class, it mislabels many "non\_white" statements as "black," indicating a taxonomy error (see Table \ref{tab:percentagemisslabelednonwhite}). The error rate of around 11\% across other classes is acceptable given the network's overall performance. Lastly, the network significantly underperforms on the "Movement" class and the "LGBTQ+" leaf, with misclassifications spread across multiple leaves in different classes (see Table \ref{tab:percentagemisslabeled}), suggesting a mismatch in definitions. The issue of mismatched definitions is a clear limitation at this stage. For cases like "black" and "non\_white," taxonomy adjustments—such as making "non\_white" the parent class of "black"—can help address misclassifications within leaves or subclasses. However, deeper issues, like the "LGBTQ+" misclassifications, may require more advanced solutions, potentially utilizing ontology matching techniques in the future.


\subsubsection{RoBERTa-Cycle-1-A}

After retraining the classifier with the new encoded filtered input, Table \ref{tab:all} shows improved accuracy for RoBERTa-Cycle-1-A and resolves the taxonomy issue for "black" and "non\_white" classes (see Table \ref{tab:selecetdclasses_cycle1A}). This performance increase is linked to the label adjustment based on the revised taxonomy. The network’s prior learning that "black" is a leaf of "non\_white" highlights the value of encoding semantic relationships into labels, enhancing label comparability and generalizability in future iterations.

\subsubsection{RoBERTa-Cycle-2}
RoBERTa-Cycle-2's class-wise performance on the dataset from \citet{vidgen-etal-2021-learning} (see Table \ref{tab:selecetdclasses_cycle2}) shows that, despite retraining, it preserves the original class definitions (e.g., "types\_of\_hate") while improving its general understanding of hate speech, as evidenced by the increase in prediction quality on the Evaluation Test Set from 0.73 to 0.84.

Although there is a slight decrease in the weighted average prediction quality from 0.89 to 0.82 on the \citet{vidgen-etal-2021-learning} dataset, this is reasonable given the complete fine-tuning. The model adapts well, correctly covering both new and old concepts, demonstrating that careful design and fine-tuning allow it to retain learned patterns while adapting to new definitions.

\section{Conclusion and Outlook}
\label{ConclusionandOutlook}
The results of this research demonstrate the feasibility of combining different hate speech taxonomies into a single, general taxonomy, which can be used to train a classifier capable of predicting a broader range of hate speech definitions. This approach reduces the need for multiple niche models, minimizing computational resources, and allows for model training without sharing sensitive data, thus addressing privacy concerns. The semantic relationships encoded in the labels also enhance generalizability for further training, aligning with current research in federated learning and continuous learning for hate speech detection.

By iteratively fine-tuning a pre-trained multi-label classifier on two distinct datasets, the research shows that a general taxonomy can improve hate speech detection, leading to higher performance in classifying general hate speech, as demonstrated on an independent evaluation test set. This work serves as proof that a general taxonomy can be used in multi-label hate speech classification, integrating diverse datasets and definitions of hate speech. It also suggests that, in the future, only trained networks need to be exchanged, not the sensitive datasets, advancing federated hate speech detection.

Looking ahead, further research is needed to explore automatic matching of taxonomies on both logical and semantic levels, including detecting mismatches based on definitions. Validation with a broader variety of hate taxonomies, and possibly the creation of a hate speech ontology, is essential. Additionally, encoding  structural knowledge through ontologies holds significant potential. Further work is needed on bias mitigation and quality assurance in the context of hate speech detection.

\section*{Limitations} 
\label{limitations}
The work has to address the following limitations. Firstly, it does not serve as a general proof that all datasets and all taxonomies can be combined into one. As seen in the work already, certain subparts of the two choose example taxonomies could not be merged. The problems seen here are similar to the problems arising and handled within the ontology matching community \citep{Shvaiko2013}, the found solutions from that field will greatly contribute to future development of the approach. Furthermore, a significant challenge is that at least the first round of training is done with possibly mislabeled data, which could lead to underperformance in the field. Similarly, the usage of algorithmically created annotations may propagate biases and underperformance, potentially even enhancing them. Lastly, the proposed iterative retraining could lead to the loss of the originally trained definitions of hate and functionality, if no countermeasures, such as more advanced subclass test sets and overfitting prevention, are conducted.

\section*{Ethical Considerations}
\label{ethicalConsideration}
Even though machine learning based applications to detect hate speech automatically online are not the solution to hate online, they are a fundamental tool in the process of combating online hate speech. this research advocated for a contextual aware human-in-the-loop strategy to counter online hate speech. The research is in the interest of society, and the public good is a central concern. The algorithmic detection of hate speech is necessary to provide a harm-free space, especially for demographic groups with special needs for protection, such as adolescents. The research is advancing the field in a more open but data-secure direction. While more diverse understandings of what constitutes hate speech is usable, the potential limitations are stated in section \ref{limitations}.

\bibliography{custom}

\appendix
\section{Appendix}

\subsection{The General Taxonomy}
\label{Generaltaxonomy}

The general taxonomy has on level 0 the classes Hate and No-hate. On level 1 it is  further broken down into Target\_of\_hate and Types\_of\_hate.   
\begin{enumerate}
\item No-hate
\item Hate 
\begin{enumerate}
\item Target\_of\_hate
\item Types\_of\_hate  
\end{enumerate}
\end{enumerate}

Target\_of\_hate is further broken down into: 
\begin{enumerate}
\item Class
\begin{enumerate}
\item Working\_class    
\end{enumerate}
\item    Immigration\_status
\begin{enumerate}
\item        Asylum\_seeker    
\item            Foreigner  
\item           Immigrants 
\item              Refugee
\end{enumerate}
\item        Movement 
\begin{enumerate}
\item         LGBTQ+  
\end{enumerate}      
\item             National\_origin 
\begin{enumerate}
\item                  China   
\item             Korea    
\item         Pakistan   
\item         Other\_N    
\item         Poland 
\item           Russian   
\end{enumerate} 
\item       Physical\_attributes 
\begin{enumerate}
\item                    Age
\begin{enumerate}
\item                     Old    
\item                   Young       
\end{enumerate} 
\item              Disability   
\item                 Gender   
\begin{enumerate}
\item       Gender\_minorities    
\begin{enumerate}
\item                  Trans   
\end{enumerate}
\item                    Man   
\item                  Women      
\end{enumerate} 
\item             Overweight   
\item            Skin\_color 
\begin{enumerate}
\item                 Black   
\item             Non\_white
\item                White 
\end{enumerate}
\end{enumerate} 
\item       Race\_Ethnicity   
\begin{enumerate}
\item                Arabs   
\item                 Asia 
\begin{enumerate}
\item                East\_A     
\item                South    
\item            South\_east   
\end{enumerate}
\item          Black\_people  
\item                Europe  
\begin{enumerate}
\item                 East\_E  
\end{enumerate}
\item               Hispanic   
\item             Indigenous 
\begin{enumerate}
\item      Aboriginal\_people  
\end{enumerate}
\item        Minority\_groups   
\item             Mixed\_race   
\item     People\_from\_Africa  
\item              Travelers  
\begin{enumerate}
\item                  Roma  
\end{enumerate}
\end{enumerate}
\item    Religion\_or\_belief 
\begin{enumerate}
\item                Hindus     
\item                  Jews  
\item               Muslims   
\item               Other\_R
\end{enumerate}
\item             Sexuality
\begin{enumerate}
\item             Sexuality     
\item              Bisexual     
\item                   Gay    
\item               Lesbian 
\end{enumerate}
\end{enumerate}

Types\_of\_hate  is further broken down into: 
\begin{enumerate} 
\item Animosity   
\item Dehumanization   
\item Derogation  
\item Support\_for\_hateful\_entities  
\item   Threatening\_language 
\end{enumerate}

\newpage
\end{document}